\documentclass[11pt,a4paper,oneside]{article}

\usepackage[abbrvbib, preprint, nohyperref]{jmlr2e}
\usepackage[a4paper,top=3.75cm,bottom=3.75cm,left=2.75cm,right=2.75cm]{geometry}
\usepackage{booktabs}
\usepackage{multirow}
\usepackage{multicol}
\usepackage{hyperref}
\usepackage{etoolbox}
\usepackage{xspace}
\usepackage[frozencache]{minted}
\usepackage{url}
\usepackage[all]{hypcap}
\usepackage[caption=false]{subfig}
\usepackage[binary-units]{siunitx}
\usepackage{researchpack}

\setminted{fontsize=\small,linenos,frame=lines,framesep=5pt,framerule=0.5pt}

\newcommand{\deeprob}{\textsc{DeeProb-kit}\xspace}
\newcommand{\spflow}{\textsc{SPFlow}\xspace}
\newcommand{\dataset}[1]{\texttt{#1}}
\newcommand{\qevi}{\ensuremath{\mathcal{Q}_\text{EVI}}}
\newcommand{\qmar}{\ensuremath{\mathcal{Q}_\text{MAR}}}
\newcommand{\qmpe}{\ensuremath{\mathcal{Q}_\text{MPE}}}
\newrobustcmd{\ubold}{\DeclareFontSeriesDefault[rm]{bf}{b}\bfseries}

\begin{document}

\title{\deeprob: a Python Library \\ for Deep Probabilistic Modelling}

\author{\name Lorenzo Loconte \email l.loconte@sms.ed.ac.uk \\
        \addr School of Informatics \\
        University of Edinburgh, UK
        \AND
        \name Gennaro Gala \email g.gala@tue.nl \\
        \addr Department of Mathematics and Computer Science \\
        Eindhoven University of Technology, The Netherlands
}

\maketitle

\begin{abstract}
\deeprob is a unified library written in Python consisting of a collection of deep probabilistic models (DPMs) that are tractable and exact representations for the modelled probability distributions.
The availability of a representative selection of DPMs in a single library makes it possible to combine them in a straightforward manner, a common practice in deep learning research nowadays.
In addition, it includes efficiently implemented learning techniques, inference routines, statistical algorithms, and provides high-quality fully-documented APIs.
The development of \deeprob will help the community to accelerate research on DPMs as well as to standardise their evaluation and better understand how they are related based on their expressivity.
\end{abstract}

\section{Introduction}

Recently, a plethora of novel probabilistic models have been introduced.
Each of them can be characterised in terms of expressiveness and tractability.
In a broad sense, a family of probabilistic models is said to be expressive if the models in it can effectively estimate a class of complex probability distributions, and its tractability corresponds to the classes of probabilistic queries that can be exactly computed in reasonable time.

In particular, Deep Probabilistic Models (DPMs) such as Variational Auto-Encoders (VAEs) \citep{kingma2014autoencoding}, Normalising Flows (NFs) \citep{papamakarios2019flows}, and Probabilistic Circuits (PCs) \citep{choi2020circuits} (also referred to as Sum-Product Networks (SPNs) \citep{poon2011spn}) gained a lot of interest recently, due to advances in deep learning techniques and an increase in available computational power.
However, these models differ for the number of probabilistic query classes they are capable of computing efficiently and exactly.
For instance, computing the probability $\Pr(\vX)$ given complete observation of variables in $\vX$ is intractable for VAEs, while being efficient for NFs and PCs.

The high fragmentation of implementations and dependencies of such models makes the usage of and integration with other deep learning techniques difficult.
That is, existing implementations of individual DPMs use several state-of-the-art libraries for high-performance computation and automatic differentiation whose combination is incompatible.
In \deeprob\footnote{The source code repository is available at \url{https://github.com/deeprob-org/deeprob-kit}} we focus on the collection of DPMs that allow evaluating (at least) complete evidence probabilistic queries exactly and efficiently by providing implementations that are easy to use, extend and combine whilst being efficient both in learning and inference phases.

\section{DeeProb-kit}

\begin{table}
    \centering
    \small
    \begin{tabular}{rlrl}
        \toprule
        \multicolumn{2}{l}{\textbf{Probabilistic Circuits}} & \multicolumn{2}{l}{\textbf{Normalizing Flows}} \\
        \midrule
        SPN            & \citet{poon2011spn}         & MAF            & \citet{papamakarios2017maf} \\
        MSPN           & \citet{molina2018mixed}     & NICE           & \citet{dinh2015nice}        \\
        XPC            & \citet{dimauro2021randompc} & RealNVP        & \citet{dinh2017realnvp}     \\
        RAT-SPN        & \citet{peharz2019ratspn}    &                &                             \\
        DGC-SPN        & \citet{wolfshaar2020dgcspn} &                &                             \\
        \bottomrule
    \end{tabular}
    \caption{Summary of Deep Probabilistic Models currently implemented in \deeprob.}
    \label{tab:models}
\end{table}

\deeprob is a library written in Python and PyTorch \citep{pytorch2019} providing several Deep Probabilistic Models (DPMs), learning techniques, inference routines and statistical algorithms.
Furthermore, extendable abstract base classes are also provided, hence permitting the introduction of novel models.
Table \ref{tab:models} lists the models currently available in the library.
In addition, we provide efficient implementations of Binary Chow-Liu Trees (Binary-CLTs) \citep{chowliu1968clt}, which are widely-used tractable probabilistic graphical models (PGMs), and Cutset Networks (CNets) \citep{rahman2014cnets} with several learning algorithms.

Existing implementations of Normalising Flows (NFs) such as in Tensorflow Probability\footnote{Code available at \url{https://github.com/tensorflow/probability}} and the original implementations from the authors\footnote{Code available at \url{https://github.com/chrischute/real-nvp} and \url{https://github.com/gpapamak/maf}} are based on incompatible deep learning libraries, i.e., either Tensorflow \citep{tensorflow2015}, Theano \citep{theano2016} or PyTorch \citep{pytorch2019}.
Furthermore, the original implementation of Randomized and Tensorized Sum-Product Networks (RAT-SPNs)\footnote{Code available at \url{https://github.com/cambridge-mlg/RAT-SPN}} is based on legacy code using Tensorflow 1.
In \deeprob, deep probabilistic models that largely benefit from GPU parallelization, such as NFs and tensorized PCs, are implemented using PyTorch.
Moreover, probabilistic models having a sparse structure such as PCs constructed via learning routines \citep{gens2013learning-structure}, still benefit from multi-core parallelization and SIMD features for their inference algorithms.

In addition, NFs are implemented in a way that allows a straightforward combination of bijective transformations, even with other DPMs.
For example, Listing \ref{lst:custom-flow} defines a custom NF having as base distribution a Gaussian RAT-SPN.
Interestingly, the combination of different DPMs may widen the landscape of existent probabilistic models, not only in terms of their expressivity, but also in terms of the required computational resources.

\begin{listing}
    \begin{minted}{python}
from deeprob.spn.models import GaussianRatSpn
from deeprob.flows.models import NormalizingFlow
from deeprob.flows.layers import CouplingLayer1d, AutoregressiveLayer

class MyFlow(NormalizingFlow):
  def __init__(self, in_features, depth, units):
    in_base = GaussianRatSpn(in_features, random_state=42)
    super().__init__(in_features, in_base=in_base, dequantize=True, logit=0.01)

    self.layers.extend([
      CouplingLayer1d(in_features, depth, units, affine=True),
      AutoregressiveLayer(in_features, depth, units, activation='tanh'),
      CouplingLayer1d(in_features, depth, units, affine=False, reverse=True)
    ])        
    \end{minted}
    \caption{Code snippet showing how a custom Normalising Flow (line 5) with a Gaussian RAT-SPN (line 7) as base distribution can be instantiated in \deeprob. Heterogeneous flow bijector layers are instantiated in sequence (lines 11-12-13). Moreover, input dequantization and logit transformations (line 8) are enabled as well.}
    \label{lst:custom-flow}
\end{listing}

\deeprob also provides versioned and high-quality documentation using Sphinx and hosted on Read the Docs\footnote{\url{https://deeprob-kit.readthedocs.io/en/latest/}}.
All the tests are automatically run on every push and pull request using GitHub Actions, and the code coverage can be visualised on Codecov\footnote{\url{https://app.codecov.io/gh/deeprob-org/deeprob-kit/}}.
Moreover, a rich set of additional scripts are provided as well.
That is, for each implemented model we include at least one example script that illustrates common usage scenarios and utility scripts to run experiments with different hyperparameters.

\section{Benchmark}

The main competitor of \deeprob in terms of Probabilistic Circuits (PCs) and Binary Chow-Liu Trees (Binary-CLTs) is \spflow \citep{molina2019spflow}, the most used library for PCs.
In order to compare the two libraries, we run benchmarks on three binary data sets with different number of samples ($N$) and variables ($V$): \dataset{msweb} ($N=29441$ and $V=294$), \dataset{bmnist} ($N=50000$ and $V=784$), and \dataset{ad} ($N=2461$ and $V=1556$).
These data sets are commonly used in literature to evaluate PCs and other probabilistic models \citep{lowd2010learning-mn-structure,larochelle2011nadist,haaren2012mnlearning}, and their relatively high number of samples and variables make them suitable for our benchmark.

The Chow-Liu algorithm \citep{chowliu1968clt} is a simple yet effective method to learn structure and parameters of a tree-shaped Bayesian Network, which takes the name of Chow-Liu Tree (CLT).
In order to fairly benchmark the implementations of PCs, we rely on the Chow-Liu algorithm.
That is, first we learn a Binary-CLT considering the entire training set split, and then compile it into an PC following the procedure in \citet{butz2020decompilation}.
This method has two advantages: first, we learn an PC using a deterministic procedure, hence allowing a fair comparison between libraries; second, the number of sum, product, and Bernoulli units of the compiled PC have the same order of magnitude.
Therefore, we also take into account the different computational effort required for different units.

We evaluate inference algorithms to compute various probabilistic query classes: complete evidence queries ($\qevi$), marginal queries ($\qmar$), and most probable explanation queries ($\qmpe$).
All the inference algorithms are executed on the training set split.
For $\qmar$ and $\qmpe$ queries, values are marginalised randomly following a Bernoulli distribution with $p=0.5$.
Specifically for Binary-CLTs, we compare the time required to learn them using the Chow-Liu algorithm offered by \spflow and \deeprob.
Finally, the last algorithm taken in consideration for PCs in our benchmark is conditional sampling, which is performed using the marginalised data set also used for $\qmar$ and $\qmpe$.

\begin{table}[H]
    \small
    \centering
    \begin{tabular}{ccS[table-format=3.2(3)]S[table-format=3.2]S[table-format=3.2(3)]S[table-format=3.2]}
        \toprule
        \multirow{2}{*}{\textbf{Dataset}} & \multirow{2}{*}{\textbf{Algorithm}} & \multicolumn{2}{c}{\spflow} & \multicolumn{2}{c}{\deeprob} \\
        & & {Time (s)} & {$\mu_{LL}$} & {Time (s)} & {$\mu_{LL}$} \\
        \midrule
        \multirow{4}{*}{\dataset{msweb}}
            & \qevi    &   0.15 \pm 0.00 & - 10.10 & 0.12 \pm 0.00 & - 10.10 \\
            & \qmar    &  62.64 \pm 1.90 & -  5.30 & 2.26 \pm 0.00 & -  5.30 \\
            & \qmpe    &  95.40 \pm 0.19 & -  6.72 & 2.09 \pm 0.01 & -  6.72 \\
            & Chow-Liu &   2.52 \pm 0.01 & {---}   & 0.04 \pm 0.00 & {---}   \\
        \midrule
        \multirow{4}{*}{\dataset{bmnist}}
            & \qevi    &   0.59 \pm 0.00 & -135.85 &  0.50 \pm 0.00 & -135.85 \\
            & \qmar    & 272.04 \pm 1.22 & - 78.92 & 10.53 \pm 0.02 & - 78.92 \\
            & \qmpe    & 430.77 \pm 1.05 & -106.56 &  9.66 \pm 0.03 & -106.56 \\
            & Chow-Liu & 160.27 \pm 0.31 & {---}   &  0.28 \pm 0.00 & {---}   \\
        \midrule
        \multirow{4}{*}{\dataset{ad}}
            & \qevi    &   0.06 \pm 0.00 & - 15.48 &  0.05 \pm 0.00 & - 15.48 \\
            & \qmar    &  25.05 \pm 0.05 & - 10.97 &  1.15 \pm 0.01 & - 10.97 \\
            & \qmpe    &  40.99 \pm 0.11 & - 12.19 &  1.01 \pm 0.00 & - 12.19 \\
            & Chow-Liu &  32.93 \pm 0.17 & {---}   &  0.56 \pm 0.00 & {---}   \\
        \bottomrule
    \end{tabular}
    \caption{Benchmark results of Binary Chow-Liu Trees (Binary-CLTs) learned on three different binary data sets through the Chow-Liu algorithm. The table shows the average time (with two standard deviations) to answer probabilistic queries and to run the Chow-Liu algorithm. For sanity check, the table also shows the average log-likelihood given by answering \qevi\ and \qmar\ queries, and the average log-likelihood of the most probable data completion.}
    \label{tab:benchmark-clts}
\end{table}

\begin{table}[H]
    \small
    \centering
    \begin{tabular}{ccS[table-format=3.2(3)]S[table-format=3.2]S[table-format=3.2(3)]S[table-format=3.2(3)]S[table-format=3.2]}
        \toprule
        \multirow{3}{*}{\textbf{Dataset}} & \multirow{3}{*}{\textbf{Algorithm}} & \multicolumn{2}{c}{\spflow} & \multicolumn{3}{c}{\deeprob} \\
        & & {Time (s)} & {$\mu_{LL}$} & \multicolumn{2}{c}{Time (s)} & {$\mu_{LL}$} \\
        & &            &            &              {$J=1$} & {$J=4$} &              \\
        \midrule
        \multirow{4}{*}{\dataset{msweb}}
            & \qevi       &  4.26 \pm 0.02 & - 10.10 &  4.27 \pm 0.01 &  1.42 \pm 0.02 & - 10.10 \\
            & \qmar       &  2.37 \pm 0.00 & -  5.30 &  2.44 \pm 0.00 &  0.93 \pm 0.01 & -  5.30 \\
            & \qmpe       &  3.87 \pm 0.07 & -  6.72 &  3.03 \pm 0.00 &  2.77 \pm 0.01 & -  6.72 \\
            & C. Sampling &  5.42 \pm 0.02 & - 10.10 &  4.13 \pm 0.01 &  1.68 \pm 0.01 & - 10.08 \\
        \midrule
        \multirow{4}{*}{\dataset{bmnist}}
            & \qevi       & 20.57 \pm 0.14 & -135.85 & 20.62 \pm 0.00 &  6.68 \pm 0.21 & -135.85 \\
            & \qmar       & 11.22 \pm 0.01 & - 78.92 & 11.35 \pm 0.00 &  4.09 \pm 0.01 & - 78.92 \\
            & \qmpe       & 20.77 \pm 0.14 & -108.66 & 14.20 \pm 0.05 & 12.79 \pm 0.07 & -108.66 \\
            & C. Sampling & 35.08 \pm 0.48 & -134.05 & 19.21 \pm 0.04 &  7.58 \pm 0.02 & -134.07 \\
        \midrule
        \multirow{4}{*}{\dataset{ad}}
            & \qevi       &  2.77 \pm 0.05 & - 15.48 &  3.13 \pm 0.01 &  2.84 \pm 0.01 & - 15.48 \\
            & \qmar       &  1.96 \pm 0.01 & - 10.97 &  2.30 \pm 0.04 &  2.66 \pm 0.01 & - 10.97 \\
            & \qmpe       &  3.28 \pm 0.11 & - 12.21 &  2.96 \pm 0.02 &  3.29 \pm 0.06 & - 12.21 \\
            & C. Sampling &  5.62 \pm 0.07 & - 15.55 &  3.94 \pm 0.01 &  4.54 \pm 0.02 & - 15.63 \\
        \bottomrule
    \end{tabular}
    \caption{Benchmark results of Probabilistic Circuits (PCs) learned on three different binary data sets. The table shows the average time (with two standard deviations) to answer probabilistic queries and to perform conditional sampling. For \deeprob we also enable multi-processing (i.e., $J=4$ parallel jobs). For sanity check, the table also shows the average log-likelihood given by answering \qevi\ and \qmar\ queries, and the average log-likelihoods of the most probable and sampled data.}
    \label{tab:benchmark-pcs}
\end{table}

Tables \ref{tab:benchmark-clts} and \ref{tab:benchmark-pcs} show our benchmark results regarding the time required to run some algorithms and to compute probabilistic queries classes on Binary-CLTs and PCs, respectively.
The results are obtained by averaging the elapsed times of 10 independent runs.
All the algorithms are executed using 32-bit floating point arithmetic on an Intel i5-4460 quad-core at \SI{3.2}{\giga\hertz} machine with \SI{8}{\gibi\byte} of RAM and Ubuntu 22.04.1 equipped with Linux kernel 5.15.0.

\section{Conclusion}

Interest in Deep Probabilistic Models (DPMs) has grown significantly in the last years.
\deeprob offers the availability of a representative selection of the most common DPMs and will improve the quality of research given the growing demand of reproducible, coherent and fair experiments.
At the time of writing, it provides several models and algorithms for Probabilistic Circuits (PCs) and Normalising Flows (NFs), as well as high-quality documentation.
The rationale of this library is that introducing new DPMs in \deeprob should always be done so as to allow straightforward combination with pre-existing ones.
Future work includes introducing new NF models and the recent advances regarding highly scalable PCs \citep{peharz2020einsum} \citep{liu2022latent-variable-distillation}.

\acks{We thank Erik Quaeghebeur, Antonio Vergari and Nicola Di Mauro for feedback on drafts of this paper.}

\bibliography{bibliography}

\begin{thebibliography}{24}
\providecommand{\natexlab}[1]{#1}
\providecommand{\url}[1]{\texttt{#1}}
\expandafter\ifx\csname urlstyle\endcsname\relax
  \providecommand{\doi}[1]{doi: #1}\else
  \providecommand{\doi}{doi: \begingroup \urlstyle{rm}\Url}\fi

\bibitem[Abadi et~al.(2015)Abadi, Agarwal, Barham, Brevdo, Chen, Citro,
  Corrado, Davis, Dean, Devin, Ghemawat, Goodfellow, Harp, Irving, Isard, Jia,
  Jozefowicz, Kaiser, Kudlur, Levenberg, Man\'{e}, Monga, Moore, Murray, Olah,
  Schuster, Shlens, Steiner, Sutskever, Talwar, Tucker, Vanhoucke, Vasudevan,
  Vi\'{e}gas, Vinyals, Warden, Wattenberg, Wicke, Yu, and
  Zheng]{tensorflow2015}
M.~Abadi, A.~Agarwal, P.~Barham, E.~Brevdo, Z.~Chen, C.~Citro, G.~S. Corrado,
  A.~Davis, J.~Dean, M.~Devin, S.~Ghemawat, I.~Goodfellow, A.~Harp, G.~Irving,
  M.~Isard, Y.~Jia, R.~Jozefowicz, L.~Kaiser, M.~Kudlur, J.~Levenberg,
  D.~Man\'{e}, R.~Monga, S.~Moore, D.~Murray, C.~Olah, M.~Schuster, J.~Shlens,
  B.~Steiner, I.~Sutskever, K.~Talwar, P.~Tucker, V.~Vanhoucke, V.~Vasudevan,
  F.~Vi\'{e}gas, O.~Vinyals, P.~Warden, M.~Wattenberg, M.~Wicke, Y.~Yu, and
  X.~Zheng.
\newblock {TensorFlow}: Large-scale machine learning on heterogeneous systems,
  2015.
\newblock URL \url{https://www.tensorflow.org/}.

\bibitem[Butz et~al.(2020)Butz, de~S.~Oliveira, and
  Peharz]{butz2020decompilation}
C.~J. Butz, J.~de~S.~Oliveira, and R.~Peharz.
\newblock Sum-product network decompilation.
\newblock In \emph{International Conference on Probabilistic Graphical Models},
  volume 138 of \emph{Proceedings of Machine Learning Research}, pages 53--64.
  {PMLR}, 2020.

\bibitem[Choi et~al.(2020)Choi, Vergari, and Van~den Broeck]{choi2020circuits}
Y.~Choi, A.~Vergari, and G.~Van~den Broeck.
\newblock Probabilistic circuits: A unifying framework for tractable
  probabilistic models.
\newblock Technical report, {UCLA Computer Science}, 2020.

\bibitem[Chow and Liu(1968)]{chowliu1968clt}
C.~K. Chow and C.~N. Liu.
\newblock Approximating discrete probability distributions with dependence
  trees.
\newblock \emph{{IEEE} Transactions on Information Theory}, 14\penalty0
  (3):\penalty0 462--467, 1968.

\bibitem[Di~Mauro et~al.(2021)Di~Mauro, Gala, Iannotta, and
  Basile]{dimauro2021randompc}
N.~Di~Mauro, G.~Gala, M.~Iannotta, and T.~M. Basile.
\newblock Random probabilistic circuits.
\newblock In \emph{Proceedings of the Thirty-Seventh Conference on Uncertainty
  in Artificial Intelligence}, volume 161 of \emph{Proceedings of Machine
  Learning Research}, pages 1682--1691. PMLR, 2021.

\bibitem[Dinh et~al.(2015)Dinh, Krueger, and Bengio]{dinh2015nice}
L.~Dinh, D.~Krueger, and Y.~Bengio.
\newblock {NICE:} non-linear independent components estimation.
\newblock In \emph{3rd International Conference on Learning Representations},
  2015.

\bibitem[Dinh et~al.(2017)Dinh, Sohl{-}Dickstein, and Bengio]{dinh2017realnvp}
L.~Dinh, J.~Sohl{-}Dickstein, and S.~Bengio.
\newblock Density estimation using real {NVP}.
\newblock In \emph{5th International Conference on Learning Representations},
  2017.

\bibitem[Gens and Pedro(2013)]{gens2013learning-structure}
R.~Gens and D.~Pedro.
\newblock Learning the structure of sum-product networks.
\newblock In \emph{Proceedings of the 30th International Conference on Machine
  Learning}, volume~28 of \emph{Proceedings of Machine Learning Research},
  pages 873--880. PMLR, 2013.

\bibitem[Haaren and Davis(2012)]{haaren2012mnlearning}
J.~V. Haaren and J.~Davis.
\newblock Markov network structure learning: {A} randomized feature generation
  approach.
\newblock In \emph{Proceedings of the Twenty-Sixth {AAAI} Conference on
  Artificial Intelligence}. {AAAI} Press, 2012.

\bibitem[Kingma and Welling(2014)]{kingma2014autoencoding}
D.~P. Kingma and M.~Welling.
\newblock Auto-encoding variational bayes.
\newblock In \emph{2nd International Conference on Learning Representations},
  2014.

\bibitem[Larochelle and Murray(2011)]{larochelle2011nadist}
H.~Larochelle and I.~Murray.
\newblock The neural autoregressive distribution estimator.
\newblock In \emph{Proceedings of the Fourteenth International Conference on
  Artificial Intelligence and Statistics}, volume~15 of \emph{{JMLR}
  Proceedings}, pages 29--37. JMLR.org, 2011.

\bibitem[Liu et~al.(2022)Liu, Zhang, and
  Broeck]{liu2022latent-variable-distillation}
A.~Liu, H.~Zhang, and G.~V.~d. Broeck.
\newblock Scaling up probabilistic circuits by latent variable distillation,
  2022.
\newblock URL \url{https://arxiv.org/abs/2210.04398}.

\bibitem[Lowd and Davis(2010)]{lowd2010learning-mn-structure}
D.~Lowd and J.~Davis.
\newblock Learning markov network structure with decision trees.
\newblock In \emph{{ICDM}}, pages 334--343. {IEEE} Computer Society, 2010.

\bibitem[Molina et~al.(2018)Molina, Vergari, {Di Mauro}, Natarajan, Esposito,
  and Kersting]{molina2018mixed}
A.~Molina, A.~Vergari, N.~{Di Mauro}, S.~Natarajan, F.~Esposito, and
  K.~Kersting.
\newblock Mixed sum-product networks: {A} deep architecture for hybrid domains.
\newblock In \emph{Proceedings of the Thirty-Second {AAAI} Conference on
  Artificial Intelligence}, pages 3828--3835. {AAAI} Press, 2018.

\bibitem[Molina et~al.(2019)Molina, Vergari, Stelzner, Peharz, Subramani, {Di
  Mauro}, Poupart, and Kersting]{molina2019spflow}
A.~Molina, A.~Vergari, K.~Stelzner, R.~Peharz, P.~Subramani, N.~{Di Mauro},
  P.~Poupart, and K.~Kersting.
\newblock {SPFlow}: An easy and extensible library for deep probabilistic
  learning using sum-product networks.
\newblock \emph{CoRR}, abs/1901.03704, 2019.

\bibitem[Papamakarios et~al.(2017)Papamakarios, Murray, and
  Pavlakou]{papamakarios2017maf}
G.~Papamakarios, I.~Murray, and T.~Pavlakou.
\newblock Masked autoregressive flow for density estimation.
\newblock In \emph{Advances in Neural Information Processing Systems 30}, pages
  2338--2347, 2017.

\bibitem[Papamakarios et~al.(2019)Papamakarios, Nalisnick, Rezende, Mohamed,
  and Lakshminarayanan]{papamakarios2019flows}
G.~Papamakarios, E.~T. Nalisnick, D.~J. Rezende, S.~Mohamed, and
  B.~Lakshminarayanan.
\newblock Normalizing flows for probabilistic modeling and inference.
\newblock \emph{CoRR}, abs/1912.02762, 2019.

\bibitem[Paszke et~al.(2019)Paszke, Gross, Massa, Lerer, Bradbury, Chanan,
  Killeen, Lin, Gimelshein, Antiga, Desmaison, Kopf, Yang, DeVito, Raison,
  Tejani, Chilamkurthy, Steiner, Fang, Bai, and Chintala]{pytorch2019}
A.~Paszke, S.~Gross, F.~Massa, A.~Lerer, J.~Bradbury, G.~Chanan, T.~Killeen,
  Z.~Lin, N.~Gimelshein, L.~Antiga, A.~Desmaison, A.~Kopf, E.~Yang, Z.~DeVito,
  M.~Raison, A.~Tejani, S.~Chilamkurthy, B.~Steiner, L.~Fang, J.~Bai, and
  S.~Chintala.
\newblock {PyTorch}: An imperative style, high-performance deep learning
  library.
\newblock In \emph{Advances in Neural Information Processing Systems 32}, pages
  8024--8035. Curran Associates, Inc., 2019.

\bibitem[Peharz et~al.(2019)Peharz, Vergari, Stelzner, Molina, Trapp, Shao,
  Kersting, and Ghahramani]{peharz2019ratspn}
R.~Peharz, A.~Vergari, K.~Stelzner, A.~Molina, M.~Trapp, X.~Shao, K.~Kersting,
  and Z.~Ghahramani.
\newblock Random sum-product networks: {A} simple and effective approach to
  probabilistic deep learning.
\newblock In \emph{Proceedings of the Thirty-Fifth Conference on Uncertainty in
  Artificial Intelligence}, volume 115 of \emph{Proceedings of Machine Learning
  Research}, pages 334--344. {AUAI} Press, 2019.

\bibitem[Peharz et~al.(2020)Peharz, Lang, Vergari, Stelzner, Molina, Trapp, den
  Broeck, Kersting, and Ghahramani]{peharz2020einsum}
R.~Peharz, S.~Lang, A.~Vergari, K.~Stelzner, A.~Molina, M.~Trapp, G.~V. den
  Broeck, K.~Kersting, and Z.~Ghahramani.
\newblock Einsum networks: Fast and scalable learning of tractable
  probabilistic circuits.
\newblock In \emph{Proceedings of the 37th International Conference on Machine
  Learning}, volume 119 of \emph{Proceedings of Machine Learning Research},
  pages 7563--7574. {PMLR}, 2020.

\bibitem[Poon and Domingos(2011)]{poon2011spn}
H.~Poon and P.~M. Domingos.
\newblock {Sum-Product Networks}: {A} new deep architecture.
\newblock In \emph{Proceedings of the Twenty-Seventh Conference on Uncertainty
  in Artificial Intelligence}, pages 337--346. {AUAI} Press, 2011.

\bibitem[Rahman et~al.(2014)Rahman, Kothalkar, and Gogate]{rahman2014cnets}
T.~Rahman, P.~Kothalkar, and V.~Gogate.
\newblock Cutset networks: {A} simple, tractable, and scalable approach for
  improving the accuracy of chow-liu trees.
\newblock In \emph{Machine Learning and Knowledge Discovery in Databases},
  volume 8725 of \emph{Lecture Notes in Computer Science}, pages 630--645.
  Springer, 2014.

\bibitem[{Theano Development Team}(2016)]{theano2016}
{Theano Development Team}.
\newblock Theano: A {Python} framework for fast computation of mathematical
  expressions.
\newblock \emph{arXiv e-prints}, abs/1605.02688, 2016.

\bibitem[van~de Wolfshaar and Pronobis(2020)]{wolfshaar2020dgcspn}
J.~van~de Wolfshaar and A.~Pronobis.
\newblock Deep generalized convolutional sum-product networks.
\newblock In \emph{International Conference on Probabilistic Graphical Models},
  volume 138 of \emph{Proceedings of Machine Learning Research}, pages
  533--544. {PMLR}, 2020.

\end{thebibliography}

\end{document}